\documentclass[10pt,twocolumn,letterpaper]{article}

\usepackage{wacv}              

\usepackage{booktabs}
\usepackage{multicol}
\usepackage{colortbl}
\usepackage{multirow}
\usepackage{pifont}
\usepackage{comment}
\usepackage[normalem]{ulem}
\usepackage{xcolor}

\usepackage{wrapfig}  
\usepackage{amsthm,amsmath,amssymb,,bm,bbm}
\usepackage{mdframed}
\usepackage{multicol}

\usepackage{mdframed}
\definecolor{theoremcolor}{rgb}{0.97, 0.97, 0.97} 
\definecolor{examplecolor}{rgb}{1, 1, 1.0}
\mdfsetup{
    innertopmargin=5pt,
    innerbottommargin=5pt,
    leftmargin=3pt,
    rightmargin=3pt,
    backgroundcolor=theoremcolor,
    linewidth=0pt,
}
\usepackage{xfrac}
\usepackage{comment}
\newmdtheoremenv[linewidth=0pt,innerleftmargin=1pt,innerrightmargin=1pt]{definition}{Definition}
\newmdtheoremenv[linewidth=0pt,innerleftmargin=2pt,innerrightmargin=2pt]{proposition}{Proposition}
\newmdtheoremenv[linewidth=0pt,innerleftmargin=0pt,innerrightmargin=0pt,backgroundcolor=examplecolor]{example}{Example}

\newmdtheoremenv{theorem}{Theorem}
\newmdtheoremenv{lemma}{Lemma}
\newmdtheoremenv{remark}{Remark}

\theoremstyle{remark}
\usepackage{floatflt}

\definecolor{wacvblue}{rgb}{0.21,0.49,0.74}
\usepackage[pagebackref,breaklinks,colorlinks,allcolors=wacvblue]{hyperref}

\def\wacvPaperID{3041} 
\def\confName{WACV}
\def\confYear{2026}

\title{Prompt-OT: An Optimal Transport Regularization Paradigm for Knowledge Preservation in Vision-Language Model Adaptation}

\author{ Xiwen Chen$^{1,2}$\thanks{\textit{These authors contributed equally to this paper}} \enspace Wenhui Zhu$^{3}$\footnotemark[1]  \enspace  Peijie Qiu$^{4}$\footnotemark[1] \enspace  Hao Wang$^{2}$ \enspace Huayu Li $^{5}$ \\
\enspace Haiyu Wu$^{5}$ \enspace Aristeidis Sotiras$^{3}$ \enspace 
  Yalin Wang$^{6}$ \enspace Abolfazl Razi$^{2}$\\
$^{1}$ Morgan Stanley,
$^{2}$ Clemson University,  
$^{3}$ Arizona State University, \\
$^{4}$ Washington University in St. Louis, 
$^{5}$ University of Arizona, 
$^{6}$ University of Notre Dame
}

\begin{document}
\maketitle

\begin{abstract}
Vision-language models (VLMs) such as CLIP demonstrate strong performance but struggle when adapted to downstream tasks. Prompt learning has emerged as an efficient and effective strategy to adapt VLMs while preserving their pre-trained knowledge. However, existing methods still lead to overfitting and degrade zero-shot generalization. To address this challenge, we propose an optimal transport (OT)-guided prompt learning framework that mitigates forgetting by preserving the structural consistency of feature distributions between pre-trained and fine-tuned models. Unlike conventional point-wise constraints, OT naturally captures cross-instance relationships and expands the feasible parameter space for prompt tuning, allowing a better trade-off between adaptation and generalization. Our approach enforces joint constraints on both vision and text representations, ensuring a holistic feature alignment. Extensive experiments on benchmark datasets demonstrate that our simple yet effective method outperforms existing prompt learning strategies in base-to-novel generalization, cross-dataset evaluation, and domain generalization, without requiring additional augmentation or ensemble techniques. The code is available at \url{https://github.com/ChongQingNoSubway/Prompt-OT}
\end{abstract}

\section{Introduction}


Foundational vision-language models (VLMs), such as CLIP \cite{radford2021learning} and ALIGN \cite{jia2021scaling}, trained on large-scale datasets to align image-text pairs, have demonstrated exceptional generalization capabilities across a variety of downstream tasks \cite{kim2022how,feng2022promptdet,luddecke2022image}. However, recent studies have shown that VLMs face challenges in maintaining generalization when adapted to downstream tasks via \textit{e.g.} fine-tuning \cite{khattak2023self}, particularly for tasks with limited samples such as few-shot learning.


\begin{figure}[!t]
    \centering
    \includegraphics[width=0.95\linewidth]{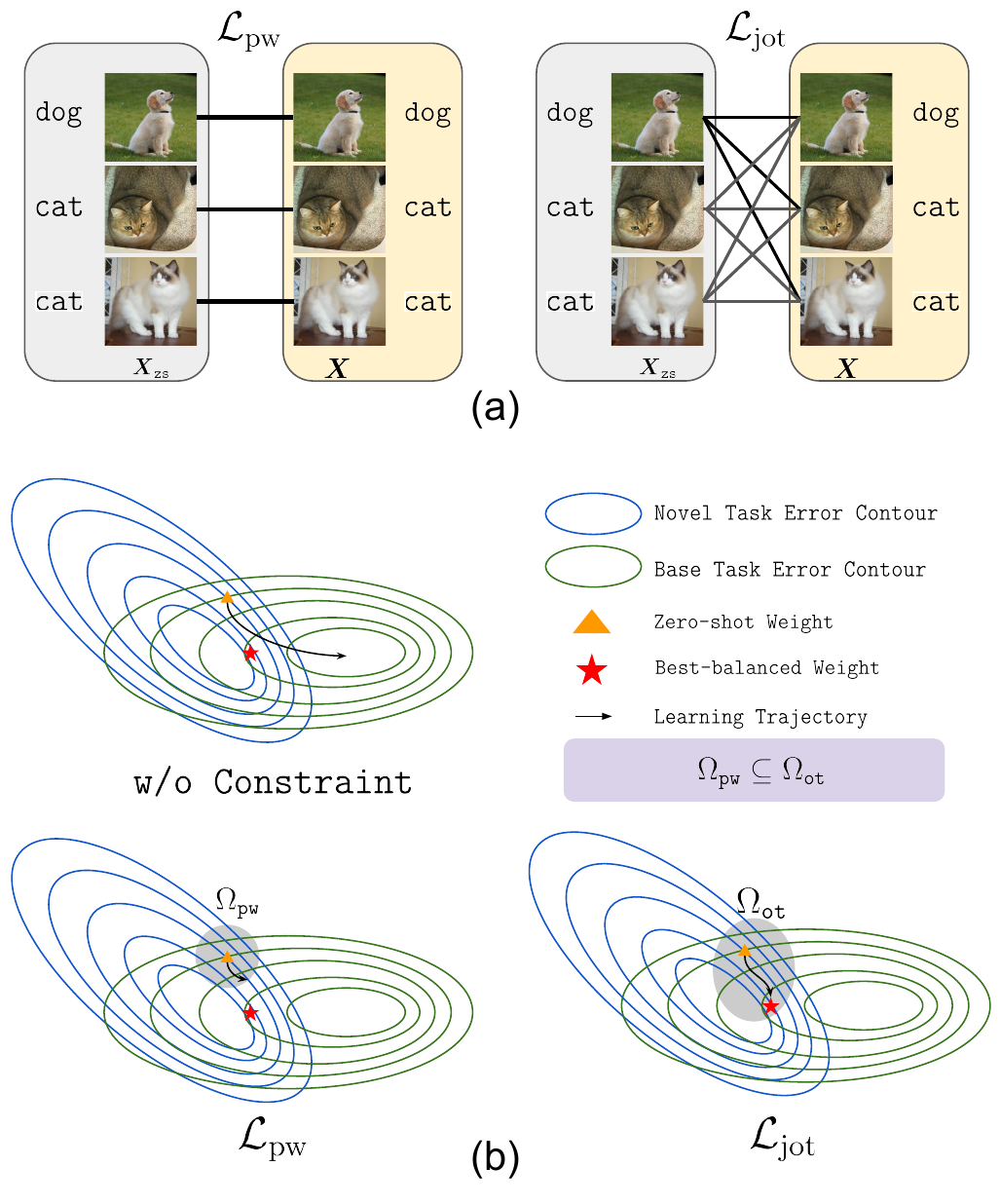}
    \caption{(a) Comparison of point-wise constraints vs. our OT-based constraints: Unlike rigid point-wise alignment, our loss captures cross-instance relationships, effectively modeling correlations both within and between classes. (b) The error contours for base and novel tasks without constraint (\textbf{top left}), with point-wise constraint (\textbf{bottom left}), and our OT-based constraints (\textbf{bottom right}). Our OT-based constraint enlarges feasible parameter domains, striking a balance between adaptation and generalization. 
}
    \label{fig:0}
\end{figure}

To tackle this issue, recent studies \cite{zhou2022learning,zhou2022conditional} have shown that introducing learnable prompt tokens while keeping the original pre-trained weights fixed is an effective strategy for fine-tuning VLMs on downstream tasks (\textit{a.k.a.} \textit{Prompt Learning}). Following this trend, many efforts have been devoted to exploring different prompt learning strategies, either through the text branch alone \cite{zhou2022learning,zhou2022conditional} or through both the text and image branches \cite{khattak2023maple,yao2024tcp}.
However, early prompt learning methods \cite{zhou2022learning,zhou2022conditional,khattak2023maple} often lead to overfitting in task-specific datasets and sacrifice zero-shot generalization. To address this limitation, more recent studies ~\cite{roy2023consistency,khattak2023self} introduce explicit regularization to maintain the feature consistency between the finetuned and the frozen pre-trained (\textit{i.e.} zero-shot) VLMs. These methods mainly employ a point/sample-wise metric (\textit{e.g.} MSE/MAE) for the feature alignment between the zero-shot and finetuned models.

Despite this, we argue that this may not be an optimal way to impose the aforementioned feature consistency due to the following reasons: \textbf{(i)} Point-wise constraints may fail to capture cross-instance relationships, limiting their ability to account for the geometry of the underlying manifold and effectively model both within-class and cross-class relations.
\textbf{(ii)} Point-wise constraints might be too restrictive\footnote{
A similar conclusion is also drawn in a well-known work \cite{kirkpatrick2017overcoming} for related tasks. We have more discussion for this method in Sec.~\ref{sec:related_work}.}, as different features and samples have different importance. This may cause the features to learn from prompted VLM to be shrunk to a narrow searching space. This limitation can hinder both adaptation and generalizability.



To address these challenges, we propose an optimal transport (OT)-guided VLMs prompt tuning method, which leverages OT to mitigate knowledge forgetting between the embedding features of the zero-shot and the finetuned VLMs. The advantages of OT-guided prompts are multi-fold. First, OT naturally captures cross-instance relationships, since the transport map establishes an interactive coupling among instances by redistributing mass based on their mutual relationships. Second, we prove that OT expands the feasible domain for learnable prompt tokens. A broader feasible domain introduces more potential local minima, provides greater flexibility for the optimizer to explore, and allows easier hyperparameter adjustment, increasing the likelihood of achieving an optimal balance between adaptation and generalization. The conceptual illustration is shown in Fig. \ref{fig:0}. 
Third, existing methods often constrain vision and text representations independently and then combine them with a weighted sum of losses. However, the vision encoder outputs features of size $B \times d$ (where $B$ and $d$ denote the batch size and the latent dimension, respectively), while the text encoder outputs features of size $C \times d$ (where $C$ denotes the number of classes). This mismatch can lead to imbalanced constraint strength across modalities. To address this, we build on the fundamental concept of an instance in CLIP, which inherently couples both vision and text. Our design therefore enforces constraints on vision and text simultaneously at the instance level, ensuring more balanced and structurally consistent multimodal representations.
Consequently, our design enforces constraints on both modalities simultaneously for each instance. It is also worth noting that, despite its simple yet effective design, our method outperforms several recent competitive approaches without relying on additional components such as ensemble methods or data augmentation techniques. In summary, our work makes the following key contributions:  
\begin{itemize}
    \item We propose an OT-based constraint scheme to mitigate knowledge forgetting, addressing the core limitations of previous constraint-based prompt learning methods. Beyond the conceptual design, we provide a \textcolor{Periwinkle}{\textit{theoretical justification}} showing that OT enlarges the feasible parameter space of learnable prompts, thereby enabling a more flexible and principled trade-off between adaptation and generalization.
   \item We propose to impose the OT constraint jointly on vision and text features of each instance. This \textcolor{Periwinkle}{\textit{balances the constraint strength across modalities}} and preserves structural consistency in multimodal representations, addressing the limitations of treating modalities independently.
    \item Empirical analysis (\textcolor{Periwinkle}{\textit{Table~\ref{tab:aba1}}}) highlights that \textcolor{Periwinkle}{\textit{Prompt-OT consistently outperforms all point-wise constraint methods with statistical significance}}.  Across three popular benchmarks (Base-to-Novel Generalization, Cross-Dataset Evaluation, and Domain Generalization), our method achieves competitive or superior performance compared to recent approaches, including PLOT~\cite{chen2023plot}, PromptSRC~\cite{khattak2023self}, TCP~\cite{yao2024tcp}, and QMaPLe~\cite{hao2025quantized}, without relying on prompt ensembles or extra augmentation. 

\end{itemize}

\section{Related Work}\label{sec:related_work}



\noindent\textbf{Adaptation to Downstream tasks.}
Two popular ways to adapt a model for downstream tasks are full fine-tuning and linear probing. However, full fine-tuning can weaken generalization, while linear probing often yields poor performance, leading to both being problematic for tasks needing accuracy and generalization, like open-vocabulary image recognition and object detection \cite{khattak2023maple}. To address this, researchers add extra learnable components without altering the original pre-trained weights, attempting to preserve both performance and generalization~\cite{zhou2022learning}. 
One representative approach is known as \textit{prompt-based methods}~\cite{zhou2022learning,zhou2022conditional,zhu2023prompt,yao2024tcp}.  
In these methods, the pre-trained model is adapted by incorporating a small number of learnable prompt tokens into the original input.
For example, CoOp \cite{zhou2022learning}, CoCoOp \cite{zhou2022conditional}, ProdGrad \cite{zhu2023prompt}, TCP \cite{yao2024tcp} introduced a set of learnable continuous vectors into the text input. Although there are some attempts to only include prompt tokens on the image branch, more recent works prefer to perform prompt learning on both branches jointly, such as MaPLe \cite{khattak2023maple} and PromptSRC \cite{khattak2023self}. Instead, adapter-based methods often add some extra layers to refine features generated by CLIP \cite{zhang2022tip,gao2024clip,yang2024mma}.

\noindent\textbf{Consistency-based Learning.} 
Instead of focusing on the design of the architecture, there are methods to enforce consistency between the trainable and pre-trained models, thereby reducing overfitting and forgetting during downstream task fine-tuning. ProGrad \cite{zhu2023prompt} aligns gradient directions with the pre-trained model, while PromptSRC \cite{khattak2023self} imposes consistency on both embeddings and logits. A similar idea in \cite{roy2023consistency} employs adapters, data augmentations, and LLM knowledge. \textit{\textcolor{Periwinkle}{Though PromptSRC is closest to our method, it only applies pairwise constraints that match average embedding distributions. We argue this is too restrictive and overlooks structural information, as discussed in Sec.~\ref{sec:Theoretical Analysis}.}} A related work \cite{kirkpatrick2017overcoming} in continual learning attempts to solve severe restriction issues through a Fisher Information-based constraint, while it requires a substantial approximation during computing and needs the corresponding weights on the frozen model, which makes it infeasible for our case, as the prompt tokens are only added to the adapted model. 
\textit{\textcolor{Periwinkle}{Additionally, previous methods typically involve terms that constrain each modality independently.}} However, during CLIP training, the feature spaces of the two modalities are inherently imbalanced: the image encoder produces a $B \times d$ feature map, while the text encoder outputs a fixed $C \times d$ representation. \textit{\textcolor{Periwinkle}{This discrepancy leads to uneven constraint strength across modalities.}} To address this, we propose imposing constraints directly on the joint representation of each instance, ensuring better alignment. A detailed analysis is provided in Sec. \ref{sec:analysis} and Table~\ref{tab:aba2}.

\noindent\textbf{Optimal Transport for VLMs.} 
Some recent studies also use OT to improve learning on CLIP, such as PLOT~\cite{chen2023plot}, Dude~\cite{nguyen2024dude}, and AWT~\cite{zhu2024awt}. PLOT employs multiple learnable text prompts and uses OT to align text and vision features. Dude leverages unbalanced OT to match class-specific and domain-shared text features (augmented by an LLM) with visual features. AWT, designed for zero-shot learning, applies various image transformations and LLM-augmented text prompts, then uses OT to measure the distance between an input image and candidate labels. \textit{\textcolor{Periwinkle}{Different from those methods, which typically use OT with augmented inputs to align vision and text features, our method applies OT as a regularization to align the vision-text joint distributions of the pre-trained and adapted models, without any need for augmentation.}}

\noindent\textbf{Augmentation for VLMs.} 
Several recent studies introduce extra augmentation to boost VLMs' adaptation. For example, \cite{khattak2023self,yao2024tcp} use more hand-crafted text templates beyond “a photo of a [\texttt{category}]” or design their prompts, while \cite{roy2023consistency,zhu2024awt} leverage LLMs for diverse semantic prompts. Others \cite{feng2023diverse} rely on Stable Diffusion \cite{rombach2022high} to generate multi-view images for better test-time tuning. These methods show that high-quality augmentation can significantly improve performance on downstream tasks. \textit{\textcolor{Periwinkle}{However, not everyone can access such data-augmentation resources. Thus, our method is designed to work without any extra augmentation, yet still delivers competitive results.}}



\section{Method}



\subsection{Preliminaries}

We first briefly present the basic pipeline of CLIP as well as the principle of \textit{Prompt Learning}.

\noindent\textbf{Vision Encoding.} An image $I$ is first divided into $M$ patches and then projected to latent space, which is denoted as $\boldsymbol{V}^0 = \{\boldsymbol{v}_1^0,\cdots,\boldsymbol{v}_{M}^0\}$. Afterward, a learnable class token $\boldsymbol{v}_{cls}^0$ is concatenated as the global representation, denoted as $[\boldsymbol{V}^0,\boldsymbol{v}_{cls}^0]$. These tokens are subsequently fed into a sequence of transformer blocks:
$
    [\boldsymbol{V}^l,\boldsymbol{v}_{cls}^l] = \texttt{Trans}^l_{\texttt{Vision}}([\boldsymbol{V}^{l-1},\boldsymbol{v}_{cls}^{l-1}]), 
$
Where $l=1,\cdots, L_I$ and $L_I$ denote the number of vision transformer blocks in the image encoder of CLIP. Afterward, the final image representation in V-L space is obtained through an additional linear projection:
$
    \boldsymbol{h}  =f_\texttt{VisionProj}(\boldsymbol{v}_{cls}^{L_I}).
$

\noindent\textbf{Text Encoding.} Each text prompt is tokenized to $N$ tokens, denoted as $\boldsymbol{T}^0 = \{\boldsymbol{t}_1^0,\cdots,\boldsymbol{t}_{N}^0\}$. Likewise, these tokens are then fed to text transformer blocks as $
    \boldsymbol{T}^l = \texttt{Trans}^l_{\texttt{Text}}(\boldsymbol{T}^{l-1}), 
$
Where $l=1,\cdots, L_T$ and $L_T$ denote the number of transformer blocks in the text encoder of CLIP.
Similarly, the final representation is obtained by the last token through a linear projection as $
    \boldsymbol{g} =f_\texttt{TextProj}(\boldsymbol{t}_{N}^{L_T}).$
    
\noindent\textbf{Prediction.} In a standard setting of zero-shot classification \cite{khattak2023maple}, the text encoder generates the hand-crafted text prompt using the common template "a photo of a [$\texttt{category}$]". Given class labels $y\in\{1,\cdots,C\}$, we can obtain the text embedding for each class by filling out the template with the corresponding class name as $(\boldsymbol{g})_y$. As mentioned earlier, we also obtained the image representation for an image as $\boldsymbol{h}$. 
The prediction can be performed as:
\begin{align}\label{eq:prediction}
p(\bar{y}|I) = \frac{\text{exp}(sim(\boldsymbol{h}, \boldsymbol{g}_y )/\tau)}{\sum_{k=1}^{C}\text{exp}(sim(\boldsymbol{h},\boldsymbol{g}_k/\tau))},
\end{align}
where $sim(\cdot,\cdot),\tau$ denote the cosine similarity and a pre-defined temperature, respectively. 

\noindent\textbf{Prompt Learning.} Now, we will delve into how to perform prompt learning, which often appends some learnable tokens to text or vision branches as part of transformer layers. Formally, we denote $\mathcal{S}_{\texttt{vision}}$ and $\mathcal{S}_{\texttt{text}}$ as the layers to integrate learnable prompt tokens for the vision encoder and text encoder, respectively. We can denote the $K_{I}$ vision prompt tokens and $K_{T}$ text prompt tokens at $(l+1)$-th block as $\boldsymbol{P}^l=\{\boldsymbol{p}^l_1,\cdots, \boldsymbol{p}^l_{K_{I}} \}$ and $\boldsymbol{Q}^l=\{\boldsymbol{q}^l_1,\cdots, \boldsymbol{q}^l_{K_{T}} \}$, respectively. The prompting for vision encoder in $l$-th layer ($l\in \mathcal{S}_{\texttt{vision}}$) can be then presented as:
\begin{align}
      [~\underline{\hspace{0.3cm}}~,\boldsymbol{V}^{l},\boldsymbol{v}_{cls}^{l}] &= \texttt{Trans}^l_{\texttt{Vision}}([\boldsymbol{P}^{l-1},\boldsymbol{V}^{l-1},\boldsymbol{v}_{cls}^{l-1}]),
\end{align}
and for text encoder in $l$-th layer ($l\in \mathcal{S}_{\texttt{text}}$) can be then presented as:
\begin{align}
    [~\underline{\hspace{0.3cm}}~, \boldsymbol{T}^l] &= \texttt{Trans}^l_{\texttt{Text}}([\boldsymbol{Q}^{l-1},\boldsymbol{T}^{l-1}]).
\end{align}
It is worth mentioning that this is known as \textit{Independent Vision-Language Prompting} (IVLP), if $\boldsymbol{P}^l$ and $\boldsymbol{Q}^l$ are defined independently. This is used in \cite{rasheed2023fine,khattak2023self}, and we will also use this baseline in our framework, following \cite{khattak2023self}. MaPLe \cite{khattak2023maple} imposes an additional linear layer to share the knowledge between $\boldsymbol{P}$ and $\boldsymbol{Q}$. 
In order to adapt classification on a downstream dataset, we can simply update the prompt tokens via cross-entropy loss $\mathcal{L}_{CE}$ between predictions (Eq. \ref{eq:prediction}) and ground truth labels.

\subsection{Vision-Text Joint Optimal Transport}
In this section, we will delve into the design of our framework. During training, this framework involves a CLIP with learnable prompts, i.e., the adapted model and a frozen pre-trained CLIP, while during inference, we only need the adapted CLIP. 

As discussed in Sec.~\ref{sec:related_work}, constraining each modality separately can result in uneven constraint strength due to inherent imbalances. To address this, we emphasize a key principle of CLIP: \textit{each instance is jointly represented in both the text and image domains}. Consequently, the two modalities should be constrained simultaneously at the instance level. Suppose we have a dataset $\mathcal{D}=\{(I_i,y_i)\}_{i=1}^n$, where $I_i,y_i$, and $n$ denote the $i$th input image, its corresponding label, and the number of samples, respectively. For each instance $i\in[n]$, the vision representation is obtained by the vision encoder as $\boldsymbol{h}^i$. Since the real text caption for the image is often unknown, we use the standard template with its category as the proxy, and accordingly, we can obtain the text representation as $\boldsymbol{g}^i$. For simplicity, we concatenate both representations as the joint representation of an instance $i$, denoted as $\boldsymbol{x}_i = \texttt{concat}(\boldsymbol{h}^i,\boldsymbol{g}^i)$. Now, we have the distribution of joint representation of the dataset $\mathcal{D}$ as $\mathcal{X}=\{\boldsymbol{x}^i\}_{i=1}^n$ by the adapted model. We also need to obtain the joint representation distribution using an additional CLIP model, i.e., the CLIP model with its original pre-trained weights, while keeping all parameters frozen. We denote it as $\boldsymbol{X}_{\texttt{zs}}=\{\boldsymbol{x}_{\texttt{zs}}^i\}_{i=1}^n$. It is worth mentioning that $\boldsymbol{x}^i$ and $\boldsymbol{x}_{\texttt{zs}}^i$ are obtained from the same input instance $(I_i,y_i)$, while with different CLIP models (i.e., one keeps updating and one is frozen). Our proposed regularization term is presented as follows:
Given $\boldsymbol{X}$ and $\boldsymbol{X}_{\texttt{zs}}$, the proposed constraint lies in Optimal Transport~\cite{villani2009optimal} is defined as:
\begin{align}\label{eq:jot}
\vspace{-0.1in}
  \mathcal{L}_{\mathrm{jot}}(\boldsymbol{X},\boldsymbol{X}_{\texttt{zs}}) =  \min_{\boldsymbol{\gamma} \in \mathcal{U}(\boldsymbol{a}, \boldsymbol{b})} \sum_{i=1}^n \sum_{j=1}^n \boldsymbol{\gamma}_{ij} \, c(\boldsymbol{x}^i, \boldsymbol{x}_{\texttt{zs}}^j),
  \vspace{-0.1in}
\end{align}
where $c(\boldsymbol{x}^i,\boldsymbol{x}_{\texttt{zs}}^j)$ denotes a cost function (e.g. linear and quadratic cost) between $\boldsymbol{x}^i$ and $\boldsymbol{x}_{\texttt{zs}}^j$ .
$\boldsymbol{\gamma}$ denotes a transport map, $\mathcal{U}(\boldsymbol{a}, \boldsymbol{b})$ denotes set of all valid transport maps,
\begin{align}\label{eq:tranportset}
\mathcal{U}(\boldsymbol{a}, \boldsymbol{b}) = \left\{ \boldsymbol{\gamma} \in \mathbb{R}^{n \times n}_{+} \ \middle| \ \boldsymbol{\gamma} \mathbf{1}_n = \boldsymbol{a}, \ \boldsymbol{\gamma}^\top \mathbf{1}_n = \boldsymbol{b} \right\}, \\ \nonumber
\boldsymbol{a} = \frac{1}{n} \mathbf{1}_n, \quad \boldsymbol{b} = \frac{1}{n} \mathbf{1}_n,
\end{align}
and \(\mathbf{1}_n\) is an \(n\)-dimensional vector of ones.



In summary, this proposed loss aims to \textit{preserve the consistency between updated joint representation and zero-shot representation while offering a larger search feasible domain of prompts and taking cross-instance from the same class or different class into consideration.} Below, we will provide a detailed analysis with theoretical justification.

\subsection{Theoretical Analysis}\label{sec:Theoretical Analysis}

Modeling cross-instance correlations can potentially characterize both within-class relations and between-class relations. 
Under the proposed OT-based constraint, each element in the transport map $\boldsymbol{\gamma}$ represents the amount of mass transported between two instances from the zero-shot representation and adapted representation. The learned transport map inherently captures cross-instance correlations because OT enforces a structured assignment, where similar instances are mapped with lower cost while maintaining the geometry of the data distribution. This process happens automatically because OT optimizes the movement of probability mass, ensuring that within-class relations remain compact while between-class relations reflect meaningful transitions, ultimately shaping a well-organized feature space. Here, we focus more on how using OT-loss can potentially enlarge the feasible parameter space, which potentially results in easier optimization and better convergence.
We first give a general definition of the loss of point-wise constraint.
     Given $\boldsymbol{X}$ and $\boldsymbol{X}_{\texttt{zs}}$, the point-wise constraint is defined as:
     \begin{align}\label{def:pw}
         \mathcal{L}_{\mathrm{pw}}(\boldsymbol{X},\boldsymbol{X}_{\texttt{zs}})= \frac{1}{n} \sum_{i=1}^nc(\boldsymbol{x}^i,\boldsymbol{x}_{\texttt{zs}}^i).
     \end{align}
We observe that it can be presented in a similar form of optimal transport:
\begin{align}\label{eq:pw-transport}
        \mathcal{L}_{\mathrm{pw}}(\boldsymbol{X},\boldsymbol{X}_{\texttt{zs}})=\sum_{i=1}^n \sum_{j=1}^n \hat{\gamma}_{ij} \, c(\boldsymbol{x}^i, \boldsymbol{x}_{\texttt{zs}}^j),
    \end{align}
    where $\forall i\neq j,\hat{\gamma}_{ij}=0$ and $\forall i= j,\hat{\gamma}_{ij}=1/n$.

\begin{lemma}\label{lemma:2}
    Let \(\boldsymbol{X}_{\texttt{zs}}\) be a zero-shot representation distribution and \(\epsilon > 0\) be a tolerance of the constraint. Suppose there exists a set \(\mathcal{X}_{\texttt{pw}}\) such that for all \(\boldsymbol{X} \in \mathcal{X}_{\texttt{pw}}\), \(\mathcal{L}_{\mathrm{pw}}(\boldsymbol{X}, \boldsymbol{X}_{\texttt{zs}}) \leq \epsilon^2\). Similarly, let there exist a set \(\mathcal{X}_{\texttt{ot}}\) such that for all \(\boldsymbol{X} \in \mathcal{X}_{\text{ot}}\), \(\mathcal{L}_{\mathrm{jot}}(\boldsymbol{X}, \boldsymbol{X}_{\texttt{zs}}) \leq \epsilon^2\). Then, it must hold that \(\mathcal{X}_{\texttt{pw}} \subseteq \mathcal{X}_{\texttt{ot}}\).
\end{lemma}
\begin{proof}
    We first note that, clearly, according to Eq.~\ref{eq:tranportset}, $\hat{\gamma}\in\mathcal{U}(\boldsymbol{a}, \boldsymbol{b})$ in Eq. \ref{eq:pw-transport} is a valid transport map. Accordingly,
    \begin{align}\label{eq:p1}
        \underbrace{\sum_{i=1}^n \sum_{j=1}^n \hat{\gamma}_{ij} \, c(\boldsymbol{x}^i, \boldsymbol{x}_{\texttt{zs}}^j)}_{\mathcal{L}_{\mathrm{pw}}} \geq  \underbrace{\min_{\boldsymbol{\gamma} \in \mathcal{U}(\boldsymbol{a}, \boldsymbol{b})} \sum_{i=1}^n \sum_{j=1}^n \boldsymbol{\gamma}_{ij} \, c(\boldsymbol{x}^i, \boldsymbol{x}_{\texttt{zs}}^j)}_{\mathcal{L}_{\mathrm{jot}}}.
    \end{align}
    This is due to $\mathcal{L}_{\mathrm{jot}}$ minimizes over all valid transport plans while $\mathcal{L}_{\mathrm{pw}}$ is applying one certain valid transport plans that not necessarily is the optimal one. Therefore, given $\boldsymbol{X}$, if $\mathcal{L}_{\text{pw}}(\boldsymbol{X}, \boldsymbol{X}_{\texttt{zs}}) \leq \epsilon^2$ holds, $\mathcal{L}_{\text{jot}}(\boldsymbol{X}, \boldsymbol{X}_{\texttt{zs}}) \leq \epsilon^2$ must hold. Hence, any $\boldsymbol{X}\in\mathcal{X}_{pw}$ satisfies the condition for $\mathcal{X}_{ot}$. 
\end{proof}

Now, we will link this theorem to the feasible parameter space of the learnable prompt tokens. First, we recap the objective function with consistency constraints:
\begin{align}
    \min_{\mathcal{P},\mathcal{Q}} \mathcal{L}_{ce} + \lambda R( \boldsymbol{X}, \boldsymbol{X}_{\texttt{zs}}),
\end{align}
where $\mathcal{L}_{ce}$ denotes the cross-entropy loss widely used in adaptation for CLIP. $\mathcal{P}=\{\boldsymbol{P}^l\}_{l\in \mathcal{S}_{\texttt{vision}}},\mathcal{Q}=\{\boldsymbol{Q}^l\}_{l\in \mathcal{S}_{\texttt{text}}}$ denote all learnable tokens from different layers. $\lambda\geq 0$ is a tuning hyperparameter. 
This objective can be considered as the generalized Lagrange function of the following optimization problem:
\begin{align}\label{eq:opt_problem}
    &\min_{\mathcal{P},\mathcal{Q}}\quad \mathcal{L}_{ce} \quad s.t.~R( \boldsymbol{X}, \boldsymbol{X}_{\texttt{zs}})-\epsilon^2\leq 0,
\end{align}
where $\epsilon^2$ is a certain constant.
Now, we have the following conclusion:
\begin{theorem}\label{theorem:1}
 Let any $\Omega_{\texttt{pw}}$ be a feasible domain of $(\mathcal{P},\mathcal{Q})$ to the optimization problem defined in Eq.~\ref{eq:opt_problem} under the point-wise constraint $\mathcal{L}_{\mathrm{pw}}$ (i.e., with $R := \mathcal{L}_{\mathrm{pw}}$). Similarly, let any $\Omega_{\texttt{ot}}$ be a feasible domain of $(\mathcal{P},\mathcal{Q})$ to Eq.~\ref{eq:opt_problem} under the optimal transport constraint $\mathcal{L}_{\mathrm{jot}}$ (i.e., with $R := \mathcal{L}_{\mathrm{jot}}$). Then, the following inclusion holds:
    \begin{equation}
        \Omega_{\texttt{pw}} \subseteq \Omega_{\texttt{jot}}.
    \end{equation}
\end{theorem}

\begin{proof} 
We define the functions $g_{\mathrm{pw}}: \Omega_{\texttt{pw}}\rightarrow \mathcal{X}_{\texttt{pw}}^*$ and $g_{\mathrm{jot}}: \Omega_{\texttt{ot}}\rightarrow \mathcal{X}_{\texttt{ot}}^*$ to represent the forward propagation of the model with dataset $\mathcal{D}$ under the point-wise constraint and the optimal transport constraint, respectively.

    Now, consider any $(\mathcal{P},\mathcal{Q}) \in \Omega_{\texttt{pw}}$. Its representation distribution $\boldsymbol{X}=g_{\mathrm{pw}}(\mathcal{P},\mathcal{Q};\mathcal{D}) $ must satisfy 
    \[
        \mathcal{L}_{\mathrm{pw}}(\boldsymbol{X}, \boldsymbol{X}_{\texttt{zs}}) \leq \epsilon^2.
    \]
    According to Lemma~\ref{lemma:2}, any $\boldsymbol{X} \in \mathcal{X}_{\texttt{pw}}^*$ also satisfies 
    \[
        \mathcal{L}_{\mathrm{jot}}(\boldsymbol{X}, \boldsymbol{X}_{\texttt{zs}}) \leq \epsilon^2.
    \]
    This implies that any $\boldsymbol{X} \in \mathcal{X}_{\texttt{pw}}^*$ also belongs to the set $\mathcal{X}_{\texttt{ot}}^*$. 

    Since for any $\boldsymbol{X} \in \mathcal{X}_{\texttt{ot}}^*$, we can find at least one corresponding $(\mathcal{P},\mathcal{Q}) \in \Omega_{\texttt{ot}}$, it follows that $\Omega_{\texttt{pw}} \subseteq \Omega_{\texttt{jot}}$.
\end{proof}

\begin{remark}
    Theorem \ref{theorem:1} immediately suggests that using optimal transport constraints allows a broader feasible parameter domain of learnable tokens $(\mathcal{P},\mathcal{Q})$. A larger feasible domain often indicates there are potentially more local minima existing and provides more pathways for the optimizer to explore, which increases the chances of finding a better solution and alleviates a heavy hyperparameter search, i.e., a uniform hyperparameter can achieve reasonable performance across different datasets. 
\end{remark}

\subsection{Training and Inference}

\noindent\textbf{Training.} Now we assemble the proposed loss in training, which is presented as \begin{align}
    \min \mathcal{L}_{ce} + \lambda \mathcal{L}_{\mathrm{jot}}( \boldsymbol{X}, \boldsymbol{X}_{\texttt{zs}}).
    \end{align}
    We employ mini-batch OT to take the mini-batch training nature, which involves computing OT distance within each batch. It is noteworthy that the mini-batch is well-bounded as studied by multiple literatures, including \cite{fatras2021minibatch,nguyen2021transportation,nguyen2021improving}.

\noindent\textbf{Inference.} Inference does not require any computation of OT and does not need the zero-shot CLIP model. Only the adapted CLIP model is used. 



\begin{table*}[!t] 
    \setlength{\tabcolsep}{4.5pt}
    \caption{\small\textbf{Comparison with state-of-the-art methods on base-to-novel generalization}. The best accuracies are bolded. HM indicates the harmonic mean. \textcolor{gray}{Prompt-OT$^\dagger$} represents the best accuracy achieved across different datasets by searching over epochs. It is not intended for direct comparison but is included here for reference. \underline{Wilcoxon signed-rank test is shown in Appendix B.}}
    \centering
    \label{table:comparision_main}
    \begin{subtable}[t]{.24\textwidth}
    \centering
    \caption{\textbf{Average}}
    \vspace{-3pt}
    \resizebox{0.99\linewidth}{!}{
    \label{table:average_acc}
    \begin{tabular}{l cc|c}
    \toprule
    & Base & Novel & HM \\
    \midrule
    CLIP & 69.34 & 74.22 & 71.70 \\
    CoOp &  {82.69} & 63.22 & 71.66 \\
    Co-CoOp & 80.47 & 71.69 & 75.83 \\
    ProGrad & 82.48 & 70.75 & 76.16 \\
    KgCoOp & 80.73 & 73.60 & 77.00 \\ 
    MaPLe & 82.28 &  {75.14} &  {78.55} \\
    PromptSRC & 84.26 & 76.10 & 79.97 \\
    ProDA & 81.56& 72.30& 76.65\\
    PLOT & 83.98& 71.72& 77.37\\ 
    TCP & 84.13& 75.36& 79.51\\ 
    QMaPLe & 83.02& 75.57& 79.12\\
     \rowcolor{gray!11}
    Prompt-OT &\textcolor{black}{\textbf{ 84.81}}& \textcolor{black}{\textbf{76.25}}& \textcolor{black}{\textbf{80.30}}\\
    \midrule
    \textcolor{gray}{Prompt-OT$^\dagger$} & \textcolor{gray}{84.82}& \textcolor{gray}{76.63}& \textcolor{gray}{80.52}\\
    \bottomrule
    \end{tabular}
    }
    \end{subtable}
    \begin{subtable}[t]{.24\textwidth}
    \centering
    \caption{ImageNet}
    \vspace{-3pt}
    \resizebox{0.99\linewidth}{!}{
    \begin{tabular}{l cc|c}
    \toprule
    & Base & Novel & HM \\
    \midrule
    CLIP & 72.43 & 68.14 & 70.22 \\
    CoOp & {76.47} & 67.88 & 71.92\\
    Co-CoOp & 75.98 & {70.43} & {73.10} \\
    ProGrad & 77.02& 66.66 &71.46\\
    KgCoOp &75.83 &69.96& 72.78\\
    MaPLe & {76.66} &  {70.54} & {73.47} \\
    PromptSRC & 77.60 & \textbf{70.73} & \textbf{74.01}\\
    ProDA & 75.40& 70.23& 72.72\\
    PLOT & 77.30& 69.87& 73.40\\ 
    TCP & 77.27& 69.87& 73.38\\ 
    QMaPLe & 76.93& \textbf{70.73}& 73.70\\
\rowcolor{gray!11}
    Prompt-OT & \textbf{77.90}& 69.83& 73.65\\
    \midrule
    \textcolor{gray}{Prompt-OT$^\dagger$} & \textcolor{gray}{77.63}& \textcolor{gray}{70.13}& \textcolor{gray}{73.69}\\
    \bottomrule
    \end{tabular}
    }
    \end{subtable}
    \begin{subtable}[t]{.24\textwidth}
    \centering
    \caption{Caltech101}
    \vspace{-3pt}
    \resizebox{0.99\linewidth}{!}{
    \begin{tabular}{l cc|c}
    \toprule
    & Base & Novel & HM \\
    \midrule
    CLIP & 96.84 & {94.00} & 95.40 \\
    CoOp & {98.00} & 89.81 & 93.73 \\
    Co-CoOp & 97.96 & 93.81 & {95.84} \\
    ProGrad & 98.02  & 93.89  & 95.91 \\ 
    KgCoOp  & 97.72  & 94.39  & 96.03 \\ 
    MaPLe & 97.74 & {94.36} &  {96.02} \\
    PromptSRC & 98.10 & 94.03 & 96.02 \\ 
    ProDA & 98.27& 93.23& 95.68\\
    PLOT & \textbf{98.53}& 92.80& 95.58\\ 
    TCP & 98.23& 94.67& 96.42\\ 
    QMaPLe & 97.97& \textbf{95.00}& \textbf{96.46}\\
    \rowcolor{gray!11}
\rowcolor{gray!11}
    Prompt-OT & 98.37& 94.50& 96.39\\
    \midrule
    \textcolor{gray}{Prompt-OT$^\dagger$} & \textcolor{gray}{98.33}& \textcolor{gray}{94.90}& \textcolor{gray}{96.59}\\
    \bottomrule

    \end{tabular}
    }
    \vspace{3pt}
    \end{subtable}
    \begin{subtable}[t]{.24\textwidth}
    \centering
    \caption{OxfordPets}
    \vspace{-3pt}
        \resizebox{0.99\linewidth}{!}{     \begin{tabular}{l cc|c}
    \toprule
    & Base & Novel & HM \\
    \midrule
    CLIP & 91.17 & 97.26 & 94.12 \\
    CoOp & 93.67 & 95.29 & 94.47 \\
    Co-CoOp & {95.20} & {97.69} & {96.43} \\
    ProGrad & 95.07 & 97.63&  96.33\\
    KgCoOp & 94.65 & 97.76 & 96.18 \\ 
    MaPLe &  {95.43} & {97.76} &  {96.58} \\
    PromptSRC & 95.33 & 97.30 & 96.30 \\ 
    ProDA & 95.43& \textbf{97.83}& \textbf{96.62}\\
    PLOT & 94.50& 96.83& 95.65\\ 
    TCP & 94.67& 97.20& 95.92\\ 
    QMaPLe & \textbf{95.67}& 97.63& 96.64\\
    \rowcolor{gray!11}
    \rowcolor{gray!11}
    Prompt-OT & 95.50& 97.03& 96.26\\
    \midrule
    \textcolor{gray}{Prompt-OT$^\dagger$} & \textcolor{gray}{95.70}& \textcolor{gray}{97.23}& \textcolor{gray}{96.46}\\
    \bottomrule
    \end{tabular}
    }
    \end{subtable}
    \\
    \begin{subtable}[t]{.24\textwidth}
    \centering
    \caption{StanfordCars}
    \vspace{-3pt}
        \resizebox{0.99\linewidth}{!}{     \begin{tabular}{l cc|c}
    \toprule
    & Base & Novel & HM \\
    \midrule
    CLIP & 63.37 &  {74.89} & 68.65 \\
    CoOp & {78.12} & 60.40 & 68.13 \\
    Co-CoOp & 70.49 & 73.59 & {72.01} \\ 
    ProGrad  & 77.68  & 68.63  & 72.88 \\ 
    KgCoOp  & 71.76  & \textbf{75.04}  & 73.36 \\ 
    MaPLe & 72.94 & 74.00 &  {73.47} \\
    PromptSRC & 78.27 & 74.97 & 76.58 \\ 
    ProDA & 74.70& 71.20& 72.91\\
    PLOT & 79.07& 74.80& 76.88\\ 
    TCP & \textbf{80.80}& 74.13& 77.32\\ 
    QMaPLe & 75.00& 73.67& 74.33\\
    \rowcolor{gray!11}
    Prompt-OT & 80.60& 74.80& \textbf{77.59}\\
    \midrule
    \textcolor{gray}{Prompt-OT$^\dagger$} & \textcolor{gray}{81.17}& \textcolor{gray}{74.70}& \textcolor{gray}{77.80}\\
    \bottomrule
    \end{tabular}
    }
    \end{subtable}
    \begin{subtable}[t]{.24\textwidth}
    \centering
    \caption{Flowers102}
    \vspace{-3pt}
        \resizebox{0.99\linewidth}{!}{     \begin{tabular}{l cc|c}
    \toprule
    & Base & Novel & HM \\
    \midrule
    CLIP & 72.08 & \textbf{77.80} & 74.83 \\
    CoOp & {97.60} & 59.67 & 74.06 \\
    Co-CoOp & 94.87 & 71.75 & {81.71} \\ 
    ProGrad  & 95.54  & 71.87  & 82.03 \\ 
    KgCoOp  & 95.00  & 74.73  & 83.65 \\ 
    MaPLe & 95.92 & 72.46 &  {82.56} \\
    PromptSRC & 98.07 & 76.50 & 85.95 \\ 
    ProDA & 97.70& 68.68& 80.66\\
    PLOT & 97.93& 73.53& 83.99\\ 
    TCP & 97.73& 75.57& 85.23\\ 
    QMaPLe & 96.43& 74.33& 83.95\\
    \rowcolor{gray!11}
    \rowcolor{gray!11}
    Prompt-OT & \textbf{98.17}& 77.03& \textbf{86.32}\\
    \midrule
    \textcolor{gray}{Prompt-OT$^\dagger$} & \textcolor{gray}{98.23}& \textcolor{gray}{77.53}& \textcolor{gray}{86.66}\\
    \bottomrule
    \end{tabular}
    }
    \end{subtable}
    \begin{subtable}[t]{.24\textwidth}
    \centering
    \caption{Food101}
    \vspace{-3pt}
        \resizebox{0.99\linewidth}{!}{     \begin{tabular}{l cc|c}
    \toprule
    & Base & Novel & HM \\
    \midrule
    CLIP & 90.10 & 91.22 & 90.66 \\
    CoOp & 88.33 & 82.26 & 85.19 \\
    Co-CoOp & {90.70} & {91.29} & {90.99} \\ 
    ProGrad & 90.37&  89.59 & 89.98 \\
    KgCoOp & 90.50 &  91.70 & 91.09 \\
    MaPLe &  {\textbf{90.71}} &  {92.05} &  {\textbf{91.38}} \\
    PromptSRC & 90.67 & 91.53 & 91.10 \\ 
    ProDA & 90.30& 88.57& 89.43\\
    PLOT & 89.80& 91.37& 90.58\\ 
    TCP & 90.57& 91.37& 90.97\\ 
    QMaPLe & 90.63&\textbf{ 92.10}& 91.36\\
    \rowcolor{gray!11}
    \rowcolor{gray!11}
    Prompt-OT & 90.67& 91.63& 91.15\\
    \midrule
    \textcolor{gray}{Prompt-OT$^\dagger$} & \textcolor{gray}{90.70}& \textcolor{gray}{92.00}& \textcolor{gray}{91.35}\\
    \bottomrule
    \end{tabular}
    }
    \end{subtable}
    \begin{subtable}[t]{.24\textwidth}
    \centering
    \caption{FGVCAircraft}
    \vspace{-3pt}
        \resizebox{0.99\linewidth}{!}{     \begin{tabular}{l cc|c}
    \toprule
    & Base & Novel & HM \\
    \midrule
    CLIP & 27.19 &  {36.29} & {31.09} \\
    CoOp & {40.44} & 22.30 & 28.75 \\
    Co-CoOp & 33.41 & 23.71 & 27.74 \\ 
    ProGrad & 40.54 & 27.57 & 32.82 \\
    KgCoOp & 36.21 & 33.55 & 34.83 \\
    MaPLe & 37.44 & 35.61 & {36.50} \\
    PromptSRC & 42.73 & \textbf{37.87} & \textbf{40.15} \\ 
    ProDA & 36.90& 34.13& 35.46\\
    PLOT & 42.13& 33.73& 37.46\\ 
    TCP & 41.97& 34.43& 37.83\\ 
    QMaPLe & 39.10& 34.90& 36.88\\
    \rowcolor{gray!11}
    \rowcolor{gray!11}
    Prompt-OT & \textbf{44.47}& 36.57& 40.12\\
    \midrule
    \textcolor{gray}{Prompt-OT$^\dagger$} & \textcolor{gray}{44.60}& \textcolor{gray}{36.67}& \textcolor{gray}{40.24}\\
    \bottomrule
    \end{tabular}
    }
    \end{subtable}
    \\
    \vspace{3pt}
    \begin{subtable}[t]{.24\textwidth}
    \centering
    \caption{SUN397}
    \vspace{-3pt}
        \resizebox{0.99\linewidth}{!}{     \begin{tabular}{l cc|c}
    \toprule
    & Base & Novel & HM \\
    \midrule
    CLIP & 69.36 & 75.35 & 72.23 \\
    CoOp & {80.60} & 65.89 & 72.51 \\
    Co-CoOp & 79.74 & {76.86} & {78.27} \\ 
    ProGrad & 81.26 & 74.17&  77.55 \\
    KgCoOp & 80.29 & 76.53 & 78.36 \\
    MaPLe & {80.82} &  {78.70} &  {79.75} \\
    PromptSRC & \textbf{82.67} & 78.47 & 80.52 \\ 
    ProDA & 78.67& 76.93& 77.79\\
    PLOT & 82.20& 73.63& 77.68\\ 
    TCP & 82.63& 78.20& 80.35\\ 
    QMaPLe & 81.33& 78.27& 79.77\\
    \rowcolor{gray!11}
    \rowcolor{gray!11}
    Prompt-OT & 82.53& \textbf{78.90}& \textbf{80.68}\\
    \midrule
    \textcolor{gray}{Prompt-OT$^\dagger$} & \textcolor{gray}{82.67}& \textcolor{gray}{78.90}& \textcolor{gray}{80.74}\\
    \bottomrule
    \end{tabular}
    }
    \end{subtable}
    \vspace{3pt}
    \begin{subtable}[t]{.24\textwidth}
    \centering
    \caption{DTD}
    \vspace{-3pt}
    \resizebox{0.99\linewidth}{!}{     \begin{tabular}{l cc|c}
    \toprule
    & Base & Novel & HM \\
    \midrule
    CLIP & 53.24 & {59.90} & 56.37 \\
    CoOp & {79.44} & 41.18 & 54.24 \\
    Co-CoOp & 77.01 & 56.00 & {64.85} \\ 
    ProGrad & 77.35&  52.35&  62.45 \\
    KgCoOp & 77.55 & 54.99 & 64.35 \\ 
    MaPLe & {80.36} & 59.18 &  {68.16} \\
    PromptSRC & \textbf{83.37} & 62.97 & 71.75 \\ 
    ProDA & 80.67& 56.48& 66.44\\
    PLOT & 81.97& 43.80& 57.09\\ 
    TCP & 82.77& 58.07& 68.25\\ 
    QMaPLe & 80.77& 57.63& 67.27\\
    \rowcolor{gray!11}
    \rowcolor{gray!11}
    Prompt-OT & \textbf{83.63}& \textbf{64.00}& \textbf{72.51}\\
    \midrule
    \textcolor{gray}{Prompt-OT$^\dagger$} & \textcolor{gray}{83.5}& \textcolor{gray}{64.27}& \textcolor{gray}{72.63}\\
    \bottomrule
    \end{tabular}
    }
    \end{subtable}
    \vspace{3pt}
    \begin{subtable}[t]{.24\textwidth}
    \centering
    \caption{EuroSAT}
    \vspace{-3pt}
    \resizebox{0.99\linewidth}{!}{     \begin{tabular}{l cc|c}
    \toprule
    & Base & Novel & HM \\
    \midrule
    CLIP & 56.48 & {64.05} & 60.03 \\
    CoOp & {92.19} & 54.74 & 68.69 \\
    Co-CoOp & 87.49 & 60.04 & {71.21} \\ 
    ProGrad & 90.11 & 60.89 & 72.67 \\ 
    KgCoOp & 85.64 & 64.34 & 73.48 \\ 
    MaPLe & {94.07} &  {73.23} & {82.35} \\
    PromptSRC & 92.90 & 73.90 & 82.32 \\ 
    ProDA & 83.90& 66.00& 73.88\\
    PLOT & 93.70& 62.67& 75.11\\ 
    TCP & 91.63& 74.73& 82.32\\ 
    QMaPLe & \textbf{94.30}& \textbf{79.47}& \textbf{86.25}\\
    \rowcolor{gray!11}
    Prompt-OT & 93.10& 75.27& 83.21\\
    \midrule
    \textcolor{gray}{Prompt-OT$^\dagger$} & \textcolor{gray}{92.77}& \textcolor{gray}{77.10}& \textcolor{gray}{84.19}\\
    \bottomrule
    \end{tabular}
    }
    \end{subtable}
    \vspace{3pt}
    \begin{subtable}[t]{.24\textwidth}
    \centering
    \caption{UCF101}
    \vspace{-3pt}
        \resizebox{0.99\linewidth}{!}{     \begin{tabular}{l cc|c}
    \toprule
    & Base & Novel & HM \\
    \midrule
    CLIP & 70.53 & {77.50} & 73.85 \\
    CoOp & {84.69} & 56.05 & 67.46 \\
    Co-CoOp & 82.33 & 73.45 & {77.64} \\ 
    ProGrad & 84.33 & 74.94 & 79.35 \\ 
    KgCoOp & 82.89 & 76.67  & 79.65 \\ 
    MaPLe & 83.00 &  {78.66} & {80.77} \\
    PromptSRC & 87.10 & 78.80 & 82.74 \\ 
    ProDA & 85.23& 71.97& 78.04\\
    PLOT & 86.60& 75.90& 80.90\\ 
    TCP & 87.13& \textbf{80.77}& \textbf{83.83}\\ 
    QMaPLe & 85.10& 77.50& 81.12\\
    \rowcolor{gray!11}
    Prompt-OT & \textbf{87.93}& 79.23& 83.36\\
    \midrule
    \textcolor{gray}{Prompt-OT$^\dagger$} & \textcolor{gray}{87.73}& \textcolor{gray}{79.50}& \textcolor{gray}{83.41}\\
    \bottomrule
    \end{tabular}
    }
    \end{subtable}
\end{table*}

\section{Experiment}
\subsection{Setups}
We follow the experimental setups of \cite{zhou2022learning,khattak2023maple,yao2024tcp} and evaluate on three tasks: Base-to-Novel Generalization, Cross-Data Evaluation, and Domain Generalization.
\textbf{For Base-to-Novel Generalization}, we train on base classes and evaluate on both base and novel classes. Following \cite{khattak2023self}, we use 11 datasets. To ensure fairness, we adopt two strategies: using the same epoch for all datasets (adhering to prior methods) and searching for the best epoch on each dataset (as a reference, only to show our method's full potential).
\textbf{For Cross-Data Evaluation}, we train on ImageNet \cite{deng2009imagenet} and test on 10 other datasets without extra fine-tuning.
\textbf{For Domain Generalization}, we again train on ImageNet \cite{deng2009imagenet} and evaluate on four out-of-distribution datasets. \underline{Please refer to Appendix A for more details.}

\begin{table*}[t]
    \caption{ Performance of PromptOT on cross-dataset evaluation and its comparison to existing methods. Here, the model is trained on the ImageNet dataset and evaluated on ten other datasets in a zero-shot setting.}
    \label{tab:cross_dataset_eval}
    \centering
    \resizebox{0.99\linewidth}{!}{
    \begin{tabular}{l  ccccccccccc}
    \toprule

    & {Caltech} & {Pets} & {Cars} & {Flowers} & {Food} & {Aircraft} & {SUN397} & {DTD} & {EuroSAT} & {UCF} & {\emph{Ave.}} \\
    \midrule
    CoOp  & 93.70 & 89.14 & 64.51 & 68.71 & 85.30 & 18.47 & 64.15 & 41.92 & {46.39} & 66.55 & 63.88 \\
    Co-CoOp & {94.43} & 90.14 & 65.32 & 71.88 & 86.06 & 22.94 & {67.36} & 45.73 & 45.37 & 68.21 & 65.74 \\
     MaPLe & 93.53 & {90.49} & {65.57} & {72.23} & {86.20} & {24.74} & 67.01 & {46.49} & {48.06} & {68.69} & {66.30} \\
     PromptSRC  & 93.60 & 90.25 & 65.70 & 70.25 & 86.15 & 23.90 & 67.10 & 46.87 & 45.50 & 68.75 & 65.81 \\
     PLOT & 92.07& 90.10& 65.70& 69.23& 86.23& 25.00& 61.67& 38.60& 47.83& 67.00& 64.34\\
     TCP & 93.97& 91.25& 64.69& 71.21& 86.69& 23.45& 67.12& 44.35& 51.45& 68.73& 66.29\\
     QCoOp & 94.07& 90.53& 65.97& 71.33& 86.23& 22.73& 66.80& 44.20& 48.23& 69.17& 65.93\\
   \midrule
    \rowcolor{gray!11}
    Prompt-OT & 94.03&90.47& 65.87& 71.27&86.43& 23.63& 67.20& 46.67& 50.57& 69.03& \textcolor{black}{\textbf{66.52}}\\ 
    \bottomrule
    \end{tabular}
    }
\end{table*}

\noindent\textbf{Baselines.} Despite compared with original CLIP \cite{radford2021learning}, we select multiple recent prompt learning methods for comparison, including: CoOp \cite{zhou2022learning}, CoCoOp \cite{zhou2022conditional}, ProDA \cite{lu2022prompt}, MaPLe \cite{khattak2023maple}, ProGrad \cite{khattak2023self}, PromptSRC \cite{khattak2023self}, PLOT \cite{chen2023plot}, TCP \cite{yao2024tcp}, and QMaPLe/QCoOP \cite{hao2025quantized}. It is worth mentioning that a portion of the methods are only designed for partial tasks, while our method does not have any obstacles to implementation for all tasks.
\textit{\textcolor{Periwinkle}{For the sake of a fair comparison and proof of concept, we will skip comparing methods that introduce substantial additional knowledge (e.g., LLMs or extra larger CLIP models for knowledge distillation) in our main experiments, such as those proposed in \cite{roy2023consistency, li2024promptkd, zhu2024awt, wu2025cascade}.}}

\noindent\textbf{Implementation details.} 
For fair comparison, we use a ViT-B/16-based CLIP model, consistent with prior work \cite{khattak2023maple,khattak2023self,yao2024tcp}. Prompts are initialized from a normal distribution, except for the text prompts in the text input, which begin with the embedding of "a photo of". We fix the learning rate at 0.005 and set $\lambda=10$ for all benchmarks. \textcolor{Periwinkle}{\textit{All reported results are the average of three runs, without any extra augmentation}}. Baseline results are taken from their respective papers. \underline{For more details, refer to Appendix~A. }

\subsection{Main Results}

\noindent\textbf{Base-to-Novel Generalization.}
As shown in Table~\ref{table:comparision_main}, our method significantly outperforms zero-shot CLIP, confirming its effectiveness. In particular, we achieve gains of 15.47\%, 2.03\%, and 8.60\% in base task accuracy, novel task accuracy, and their harmonic mean, respectively. More importantly, when averaged over 11 datasets, our approach surpasses all recent baselines. Specifically, it achieves 84.81\% (base task accuracy), 76.26\% (novel task accuracy), and 80.30\% (harmonic mean), which is an improvement of 0.55\%, 0.15\%, and 0.33\% over the previous state-of-the-art, PromptSRC.

    \noindent\textbf{Cross-Dataset Evaluation.} Table~\ref{tab:cross_dataset_eval} shows that our method attains an average accuracy of 66.52\% across 11 datasets, outperforming all baselines and exceeding PromptSRC by 0.62\%.
\begin{table}[t]
\centering
\caption{Performance on domain generalization. These baseline models are trained on ImageNet and evaluated on four out-of-distribution ImageNet datasets.}
\resizebox{0.99\linewidth}{!}{
\begin{tabular}{l ccccc}
    \toprule
    
      & ImNetV2 & ImNetS & ImNetA & ImNetR & \textit{Ave.} \\
    \midrule
    CLIP  & 60.83 & {46.15} & 47.77 & {73.96} & 57.17 \\
    UPT  & 64.35 & 48.66 & 50.66& 76.24 &59.98 \\
    CoOp  & 64.20 & 47.99  & 49.71  & 75.21  & 59.28\\
    Co-CoOp  & {64.07} & 48.75 & 50.63 & 76.18  & 59.90\\
    ProGrad  & 64.73  & 47.61 & 49.39 & 74.58 & 59.07 \\ 
    KgCoOp  & 64.10 & 48.97 & 50.69&  76.70 & 60.11 \\ 
    MaPLe  & {64.07} & {49.15} & {50.90}  & {76.98} & 60.26\\
    
    PromptSRC  & 64.35 & 49.55 & 50.90 & 77.80 & 60.65\\
    TCP & 64.60  & 49.50  & 51.20  & 76.73  & 60.51 \\
    QCoOp  & 63.87  & 48.93  & 51.10  & 76.90  & 60.20 \\
    \midrule
    \rowcolor{gray!11} 
     Prompt-OT  & 64.35 & 49.40 & 51.63& 77.40 & \textcolor{black}{\textbf{60.70}}\\
    \bottomrule
    \end{tabular}
}
    \label{tab:domain_generalization}
\end{table}
    
\noindent\textbf{Domain Generalization.} 
As shown in Table~\ref{tab:domain_generalization}, we observe that our method achieves the highest average accuracy on four out-of-distribution datasets. 

\noindent Overall, our method demonstrates robust performance across three key evaluations. In the base-to-novel generalization setup, it substantially outperforms zero-shot CLIP and surpasses the previous state-of-the-art in base accuracy, novel accuracy, and harmonic mean. In cross-dataset evaluation and in domain generalization, our method obtains the best overall accuracy without relying on prompt ensembles or text augmentation, underscoring both its effectiveness and simplicity.

\section{Analysis}\label{sec:analysis}

In this section, we will delve into a more detailed analysis of our proposed method. All experiments will be conducted consistently with the setups of base-to-novel generalization.
For fairness, we exclude all other components and only apply regularization on the IVLP.


\paragraph{Ours VS Point-wise Constraints.}
To illustrate the superiority of the proposed method over point-wise constraints, we evaluate it against several baselines:  
(i) naive $L_2$ constraints between $\boldsymbol{X}$ and $\boldsymbol{X}_{\texttt{zs}}$ with the same regularization strength $\lambda$,  
(ii) \textbf{Adaptor-Cos}~\cite{roy2023consistency}, which introduces adapters at the end of the adapted encoders and employs cosine similarity as the point-wise constraint between representations, 
(iii) \textbf{SRC}~\cite{khattak2023self}, which applies $L_1$ losses for vision and text representations separately and a KL-divergence between prediction probabilities for each instance. 
\begin{table}
\centering
\caption{Comparison of performance with point-wise constraints. Reported $p$-values are obtained from Wilcoxon signed-rank tests comparing the baseline and our method.}
\label{tab:aba1}
\resizebox{0.99\linewidth}{!}{%
\begin{tabular}{cccccc}\toprule
\multicolumn{2}{c}{Reg. Method} & \textbf{Base} & \textbf{Novel} & \textbf{HM} & $p<0.05$ \\ \midrule
\multicolumn{2}{c}{w/o Constraints} & 84.20 & 71.21 & 77.16 & \ding{51} \\ \midrule

\multirow{3}{*}{\rotatebox{90}{\textbf{Point-wise}}} 
  & $L_2$ between $\boldsymbol{X}, \boldsymbol{X}_{\texttt{zs}}$ & 84.20 & 74.68 & 79.16 & \ding{51} \\ \cline{2-6}
  & \begin{tabular}[c]{@{}c@{}}\textbf{Adaptor-Cos}\\ \textit{From CoPrompt~\cite{roy2023consistency}}\end{tabular} 
    & 83.12 & 75.25 & 78.99 & \ding{51} \\ \cline{2-6}
  & \begin{tabular}[c]{@{}c@{}}\textbf{SRC}\\ \textit{From PromptSRC~\cite{khattak2023self}}\end{tabular}         
    & 84.41 & 74.89 & 79.37 & \ding{51} \\ \midrule
\multicolumn{2}{c}{\textbf{Separate OT}} & 84.50 & 75.48 & 79.74 & \ding{51} \\ \midrule
\rowcolor{gray!11}\multicolumn{2}{c}{\textbf{Ours}} & 84.81 & 76.25 & \textbf{80.30} & -- \\ \bottomrule
\end{tabular}%
}
\end{table}
(iv) \textbf{Separate OT} (\underline{see Appendix A}), which applies OT-loss for vision and text representations separately. 
\textcolor{Periwinkle}{\textit{For a fair comparison, we exclude all additional augmentations, such as ensemble techniques and LLM-based augmentation.}} Table~\ref{tab:aba1} reports both the performance metrics and the corresponding statistical test results.
Our primary observation is that all these methods achieve a better balance between novel and base tasks than the model without any constraints, with a minimum improvement of 1.83\% in harmonic mean accuracy. More importantly, the OT-based methods outperform all recent approaches based on point-wise constraints. Even without joint representation, applying OT for two domains separately can outperform all other point-wise methods. Additionally, compared to SRC, the previous state-of-the-art method, our proposed method achieved statistically significant gains of 0.4\%, 1.35\%, and 0.93\% in base task accuracy, novel task accuracy, and harmonic mean accuracy, respectively.
Compared to $L_2$ constraints, we observe that with the same regularization strength $\lambda$, $L_2$ constraints, while preserving good generalizability to novel tasks, fail to yield any improvements on the base task. This suggests that $L_2$ constraints may be overly restrictive for model adaptation. In contrast, our proposed approach effectively balances performance across both the base and novel tasks, demonstrating that modeling cross-instance correlation while allowing the model greater flexibility in updating leads to an optimal trade-off, as expected in Fig.~\ref{fig:0}.


\paragraph{Effectiveness of Joint Representation.} 
Here, we investigate alternative approaches to imposing constraints between zero-shot and adapted representations. Specifically, we examine four different strategies: (i) imposing the constraint solely on the text representation, (ii) imposing the constraint solely on the vision representation, (iii) applying constraints separately to text and vision representation distributions (i.e., Separate OT), and (iv) our proposed joint representation, which considers an instance includes both vision and text representations.
\begin{table}[]
\centering
\caption{Ablation study on different modalities for imposing constraints.}
\label{tab:aba2}
\resizebox{0.9\linewidth}{!}{%
\addtolength{\tabcolsep}{-0.15em}
\begin{tabular}{cccccc}\toprule
Text & Vision & Scheme & \textbf{Base} & \textbf{Novel} & \textbf{HM} \\ \midrule
\ding{55} & \ding{55} & - & 84.20 & 71.21 & 77.16 \\
\ding{51} & \ding{55} & - & 84.92 & 71.75 & 77.78 \\
\ding{55} & \ding{51} & - & 84.12 & 72.94 & 78.12 \\
\ding{51} & \ding{51} & Separate & 84.50 & 75.48 & 79.74 \\ \midrule
\rowcolor{gray!11}\ding{51} & \ding{51} & Joint & 84.81 & 76.25 & \textbf{80.30} \\ \bottomrule
\end{tabular}%
}
\end{table}

As shown in Table \ref{tab:aba2}, applying constraints to both modalities yields a significantly greater improvement (2.58\% and 3.14\% in HM) compared to imposing the constraint on a single modality (0.62\% and 0.96\% in HM). Moreover, our proposed joint representation yields greater improvements compared to applying constraints separately on vision and text distributions, which potentially results in unbalanced constraint strength. Our method leads to the highest increase in HM (3.14\%), suggesting that it provides the most effective balance between minimizing errors in both novel and base classes.

\begin{figure}[!t]
        \centering
        \includegraphics[width=0.99\linewidth]{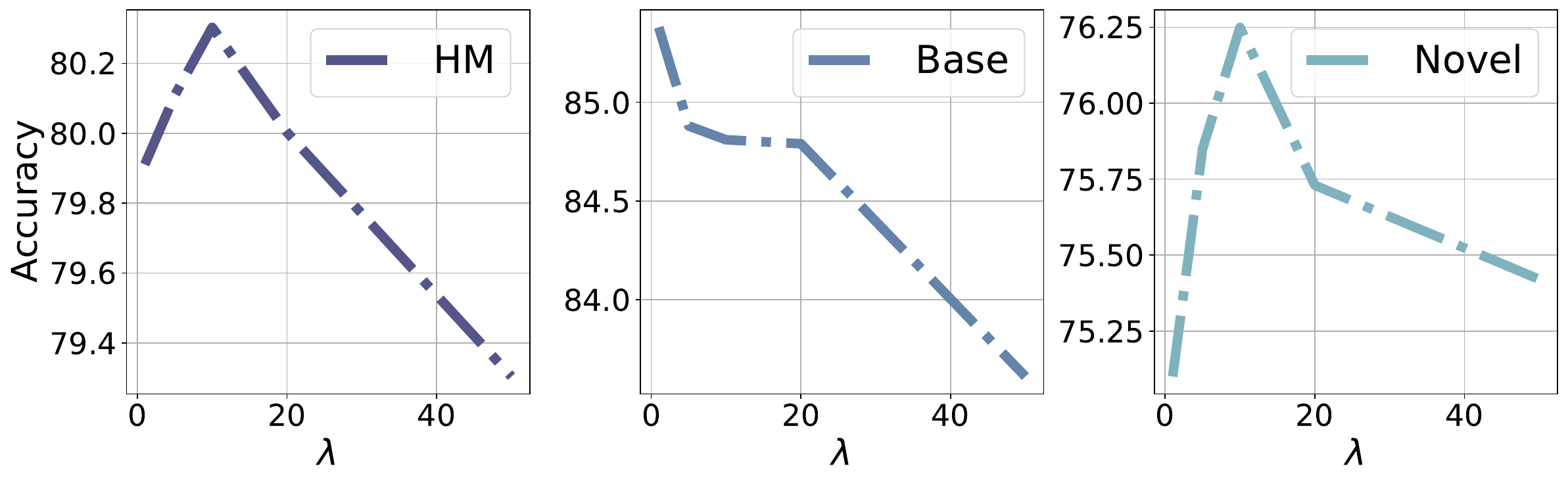}
        \caption{The effectiveness of $\lambda$ on base-to-novel generalization tasks averaged over 11 datasets.}
        \label{fig:lamda}
\end{figure}

\paragraph{Sensitivity of $\lambda$.} 
We further conduct a study to evaluate the impact of different values of $\lambda$ on adaptation. As shown in Fig.~\ref{fig:lamda}, as $\lambda$ increases, the constraints become stronger, leading to a decline in adaptation to the base class. Meanwhile, the accuracy on the novel class initially increases before decreasing, indicating that the zero-shot weights may not be optimal for both novel and base tasks. This suggests that some relaxation is necessary to achieve a well-balanced trade-off. Excessively strong constraints may prevent the model from reaching the low-error regions of both tasks in the optimization landscape.
Empirically, we find that $\lambda=10$ achieves a good balance between generalization and adaptation.

\paragraph{Efficiency.} Our method introduces only a minor computational overhead during training, as shown in Table \ref{tab:overhead}, incurring approximately a 1.8\% increase in runtime with negligible additional GPU usage.

\begin{table}[h]
   \caption{Comparison of Computational Cost.}
    \label{tab:overhead}
    \centering
    \resizebox{0.8\linewidth}{!}{%
        \begin{tabular}{lcc}
            \toprule
            & Runtime & GPU \\ 
            \midrule
            w/o Constraint & 0.0385s/step & 3948MB \\
            \rowcolor{gray!11}\textbf{Ours} & 0.0392s/step & 3950MB \\
            \bottomrule
        \end{tabular}
    }
\end{table}




\section{Conclusion}
In this work, we proposed an Optimal Transport (OT) Regularization framework for vision-language prompt learning to mitigate forgetting and enhance adaptability. Unlike rigid point-wise constraints, our method uses OT to model cross-instance relationships and preserve the embedding structure between pre-trained and fine-tuned models. Theoretical analysis shows that OT constraints offer a broader feasible parameter space, enabling more robust prompt learning. Extensive experiments demonstrate that our approach outperforms state-of-the-art methods across various generalization settings. Future directions include incorporating adaptive regularization strength to improve trade-offs dynamically and exploring efficient transport solvers for broader applications.

{
    \small
    \bibliographystyle{ieeenat_fullname}
    \bibliography{main}

\begin{thebibliography}{46}
\providecommand{\natexlab}[1]{#1}
\providecommand{\url}[1]{\texttt{#1}}
\expandafter\ifx\csname urlstyle\endcsname\relax
  \providecommand{\doi}[1]{doi: #1}\else
  \providecommand{\doi}{doi: \begingroup \urlstyle{rm}\Url}\fi

\bibitem[Bossard et~al.(2014)Bossard, Guillaumin, and Gool]{bossard2014food}
Lukas Bossard, Matthieu Guillaumin, and Luc~Van Gool.
\newblock Food-101--mining discriminative components with random forests.
\newblock In \emph{ECCV}, pages 446--461. Springer, 2014.

\bibitem[Chen et~al.(2023)Chen, Yao, Song, Li, Rao, and Zhang]{chen2023plot}
Guangyi Chen, Weiran Yao, Xiangchen Song, Xinyue Li, Yongming Rao, and Kun Zhang.
\newblock {PLOT}: Prompt learning with optimal transport for vision-language models.
\newblock In \emph{The Eleventh International Conference on Learning Representations}, 2023.

\bibitem[Cimpoi et~al.(2014)Cimpoi, Maji, Kokkinos, Mohamed, and Vedaldi]{cimpoi2014describing}
Mircea Cimpoi, Subhransu Maji, Iasonas Kokkinos, Sammy Mohamed, and Andrea Vedaldi.
\newblock Describing textures in the wild.
\newblock In \emph{CVPR}, pages 3606--3613, 2014.

\bibitem[Dem{\v{s}}ar(2006)]{demvsar2006statistical}
Janez Dem{\v{s}}ar.
\newblock Statistical comparisons of classifiers over multiple data sets.
\newblock \emph{The Journal of Machine learning research}, 7:\penalty0 1--30, 2006.

\bibitem[Deng et~al.(2009)Deng, Dong, Socher, Li, Li, and Fei-Fei]{deng2009imagenet}
Jia Deng, Wei Dong, Richard Socher, Li-Jia Li, Kai Li, and Li Fei-Fei.
\newblock Imagenet: A large-scale hierarchical image database.
\newblock In \emph{CVPR}, pages 248--255. Ieee, 2009.

\bibitem[Fatras et~al.(2021)Fatras, Zine, Majewski, Flamary, Gribonval, and Courty]{fatras2021minibatch}
Kilian Fatras, Younes Zine, Szymon Majewski, R{\'e}mi Flamary, R{\'e}mi Gribonval, and Nicolas Courty.
\newblock Minibatch optimal transport distances; analysis and applications.
\newblock \emph{arXiv preprint arXiv:2101.01792}, 2021.

\bibitem[Fei-Fei et~al.(2004)Fei-Fei, Fergus, and Perona]{fei2004learning}
Li Fei-Fei, Rob Fergus, and Pietro Perona.
\newblock Learning generative visual models from few training examples: An incremental bayesian approach tested on 101 object categories.
\newblock In \emph{CVPR Workshop}, pages 178--178. IEEE, 2004.

\bibitem[Feng et~al.(2022)Feng, Zhong, Jie, Chu, Ren, Wei, Xie, and Ma]{feng2022promptdet}
Chengjian Feng, Yujie Zhong, Zequn Jie, Xiangxiang Chu, Haibing Ren, Xiaolin Wei, Weidi Xie, and Lin Ma.
\newblock Promptdet: Towards open-vocabulary detection using uncurated images.
\newblock In \emph{ECCV}, 2022.

\bibitem[Feng et~al.(2023)Feng, Yu, Liu, Khan, and Zuo]{feng2023diverse}
Chun-Mei Feng, Kai Yu, Yong Liu, Salman Khan, and Wangmeng Zuo.
\newblock Diverse data augmentation with diffusions for effective test-time prompt tuning.
\newblock In \emph{Proceedings of the IEEE/CVF International Conference on Computer Vision}, pages 2704--2714, 2023.

\bibitem[Gao et~al.(2024)Gao, Geng, Zhang, Ma, Fang, Zhang, Li, and Qiao]{gao2024clip}
Peng Gao, Shijie Geng, Renrui Zhang, Teli Ma, Rongyao Fang, Yongfeng Zhang, Hongsheng Li, and Yu Qiao.
\newblock Clip-adapter: Better vision-language models with feature adapters.
\newblock \emph{International Journal of Computer Vision}, 132\penalty0 (2):\penalty0 581--595, 2024.

\bibitem[Hao et~al.(2025)Hao, Ding, Feng, Yang, Chen, and Ding]{hao2025quantized}
Tianxiang Hao, Xiaohan Ding, Juexiao Feng, Yuhong Yang, Hui Chen, and Guiguang Ding.
\newblock Quantized prompt for efficient generalization of vision-language models.
\newblock In \emph{European Conference on Computer Vision}, pages 54--73. Springer, 2025.

\bibitem[Helber et~al.(2019)Helber, Bischke, Dengel, and Borth]{helber2019eurosat}
Patrick Helber, Benjamin Bischke, Andreas Dengel, and Damian Borth.
\newblock Eurosat: A novel dataset and deep learning benchmark for land use and land cover classification.
\newblock \emph{J-STARS}, 12\penalty0 (7):\penalty0 2217--2226, 2019.

\bibitem[Hendrycks et~al.(2021{\natexlab{a}})Hendrycks, Basart, Mu, Kadavath, Wang, Dorundo, Desai, Zhu, Parajuli, Guo, et~al.]{hendrycks2021many}
Dan Hendrycks, Steven Basart, Norman Mu, Saurav Kadavath, Frank Wang, Evan Dorundo, Rahul Desai, Tyler Zhu, Samyak Parajuli, Mike Guo, et~al.
\newblock The many faces of robustness: A critical analysis of out-of-distribution generalization.
\newblock In \emph{ICCV}, pages 8340--8349, 2021{\natexlab{a}}.

\bibitem[Hendrycks et~al.(2021{\natexlab{b}})Hendrycks, Zhao, Basart, Steinhardt, and Song]{hendrycks2021natural}
Dan Hendrycks, Kevin Zhao, Steven Basart, Jacob Steinhardt, and Dawn Song.
\newblock Natural adversarial examples.
\newblock In \emph{CVPR}, pages 15262--15271, 2021{\natexlab{b}}.

\bibitem[Jia et~al.(2021)Jia, Yang, Xia, Chen, Parekh, Pham, Le, Sung, Li, and Duerig]{jia2021scaling}
Chao Jia, Yinfei Yang, Ye Xia, Yi-Ting Chen, Zarana Parekh, Hieu Pham, Quoc Le, Yun-Hsuan Sung, Zhen Li, and Tom Duerig.
\newblock Scaling up visual and vision-language representation learning with noisy text supervision.
\newblock In \emph{ICML}, pages 4904--4916. PMLR, 2021.

\bibitem[Khattak et~al.(2023{\natexlab{a}})Khattak, Rasheed, Maaz, Khan, and Khan]{khattak2023maple}
Muhammad~Uzair Khattak, Hanoona Rasheed, Muhammad Maaz, Salman Khan, and Fahad~Shahbaz Khan.
\newblock Maple: Multi-modal prompt learning.
\newblock In \emph{CVPR}, pages 19113--19122, 2023{\natexlab{a}}.

\bibitem[Khattak et~al.(2023{\natexlab{b}})Khattak, Wasim, Naseer, Khan, Yang, and Khan]{khattak2023self}
Muhammad~Uzair Khattak, Syed~Talal Wasim, Muzammal Naseer, Salman Khan, Ming-Hsuan Yang, and Fahad~Shahbaz Khan.
\newblock Self-regulating prompts: Foundational model adaptation without forgetting.
\newblock In \emph{Proceedings of the IEEE/CVF International Conference on Computer Vision}, pages 15190--15200, 2023{\natexlab{b}}.

\bibitem[Kim et~al.(2022)Kim, Laskin, Mordatch, and Pathak]{kim2022how}
Konwoo Kim, Michael Laskin, Igor Mordatch, and Deepak Pathak.
\newblock How to adapt your large-scale vision-and-language model, 2022.

\bibitem[Kirkpatrick et~al.(2017)Kirkpatrick, Pascanu, Rabinowitz, Veness, Desjardins, Rusu, Milan, Quan, Ramalho, Grabska-Barwinska, et~al.]{kirkpatrick2017overcoming}
James Kirkpatrick, Razvan Pascanu, Neil Rabinowitz, Joel Veness, Guillaume Desjardins, Andrei~A Rusu, Kieran Milan, John Quan, Tiago Ramalho, Agnieszka Grabska-Barwinska, et~al.
\newblock Overcoming catastrophic forgetting in neural networks.
\newblock \emph{Proceedings of the national academy of sciences}, 114\penalty0 (13):\penalty0 3521--3526, 2017.

\bibitem[Krause et~al.(2013)Krause, Stark, Deng, and Fei-Fei]{krause20133d}
Jonathan Krause, Michael Stark, Jia Deng, and Li Fei-Fei.
\newblock 3d object representations for fine-grained categorization.
\newblock In \emph{ICCV}, pages 554--561, 2013.

\bibitem[Li et~al.(2024)Li, Li, Fu, Zhang, Wang, Chen, and Yang]{li2024promptkd}
Zheng Li, Xiang Li, Xinyi Fu, Xin Zhang, Weiqiang Wang, Shuo Chen, and Jian Yang.
\newblock Promptkd: Unsupervised prompt distillation for vision-language models.
\newblock In \emph{Proceedings of the IEEE/CVF Conference on Computer Vision and Pattern Recognition}, pages 26617--26626, 2024.

\bibitem[Lu et~al.(2022)Lu, Liu, Zhang, Liu, and Tian]{lu2022prompt}
Yuning Lu, Jianzhuang Liu, Yonggang Zhang, Yajing Liu, and Xinmei Tian.
\newblock Prompt distribution learning.
\newblock In \emph{CVPR}, pages 5206--5215, 2022.

\bibitem[L{\"u}ddecke and Ecker(2022)]{luddecke2022image}
Timo L{\"u}ddecke and Alexander Ecker.
\newblock Image segmentation using text and image prompts.
\newblock In \emph{CVPR}, pages 7086--7096, 2022.

\bibitem[Maji et~al.(2013)Maji, Rahtu, Kannala, Blaschko, and Vedaldi]{maji2013fine}
Subhransu Maji, Esa Rahtu, Juho Kannala, Matthew Blaschko, and Andrea Vedaldi.
\newblock Fine-grained visual classification of aircraft.
\newblock \emph{arXiv preprint arXiv:1306.5151}, 2013.

\bibitem[Nguyen et~al.(2024)Nguyen, Le, Nguyen, Diep, Nguyen, Duong-Tran, Peters, Shen, Niepert, and Sonntag]{nguyen2024dude}
Duy~MH Nguyen, An~T Le, Trung~Q Nguyen, Nghiem~T Diep, Tai Nguyen, Duy Duong-Tran, Jan Peters, Li Shen, Mathias Niepert, and Daniel Sonntag.
\newblock Dude: Dual distribution-aware context prompt learning for large vision-language model.
\newblock \emph{arXiv preprint arXiv:2407.04489}, 2024.

\bibitem[Nguyen et~al.(2022{\natexlab{a}})Nguyen, Nguyen, Nguyen, Pham, Bui, Phung, Le, and Ho]{nguyen2021transportation}
Khai Nguyen, Dang Nguyen, Quoc Nguyen, Tung Pham, Hung Bui, Dinh Phung, Trung Le, and Nhat Ho.
\newblock On transportation of mini-batches: A hierarchical approach.
\newblock In \emph{Proceedings of the 39th International Conference on Machine Learning}, 2022{\natexlab{a}}.

\bibitem[Nguyen et~al.(2022{\natexlab{b}})Nguyen, Nguyen, Pham, and Ho]{nguyen2021improving}
Khai Nguyen, Dang Nguyen, Tung Pham, and Nhat Ho.
\newblock Improving mini-batch optimal transport via partial transportation.
\newblock In \emph{Proceedings of the 39th International Conference on Machine Learning}, 2022{\natexlab{b}}.

\bibitem[Nilsback and Zisserman(2008)]{nilsback2008automated}
Maria-Elena Nilsback and Andrew Zisserman.
\newblock Automated flower classification over a large number of classes.
\newblock In \emph{ICVGIP}, pages 722--729. IEEE, 2008.

\bibitem[Parkhi et~al.(2012)Parkhi, Vedaldi, Zisserman, and Jawahar]{parkhi2012cats}
Omkar~M Parkhi, Andrea Vedaldi, Andrew Zisserman, and CV Jawahar.
\newblock Cats and dogs.
\newblock In \emph{CVPR}, pages 3498--3505. IEEE, 2012.

\bibitem[Radford et~al.(2021)Radford, Kim, Hallacy, Ramesh, Goh, Agarwal, Sastry, Askell, Mishkin, Clark, et~al.]{radford2021learning}
Alec Radford, Jong~Wook Kim, Chris Hallacy, Aditya Ramesh, Gabriel Goh, Sandhini Agarwal, Girish Sastry, Amanda Askell, Pamela Mishkin, Jack Clark, et~al.
\newblock Learning transferable visual models from natural language supervision.
\newblock In \emph{ICML}, pages 8748--8763. PMLR, 2021.

\bibitem[Rasheed et~al.(2023)Rasheed, Khattak, Maaz, Khan, and Khan]{rasheed2023fine}
Hanoona Rasheed, Muhammad~Uzair Khattak, Muhammad Maaz, Salman Khan, and Fahad~Shahbaz Khan.
\newblock Fine-tuned clip models are efficient video learners.
\newblock In \emph{CVPR}, pages 6545--6554, 2023.

\bibitem[Recht et~al.(2019)Recht, Roelofs, Schmidt, and Shankar]{recht2019imagenet}
Benjamin Recht, Rebecca Roelofs, Ludwig Schmidt, and Vaishaal Shankar.
\newblock Do imagenet classifiers generalize to imagenet?
\newblock In \emph{ICML}, pages 5389--5400. PMLR, 2019.

\bibitem[Rombach et~al.(2022)Rombach, Blattmann, Lorenz, Esser, and Ommer]{rombach2022high}
Robin Rombach, Andreas Blattmann, Dominik Lorenz, Patrick Esser, and Bj{\"o}rn Ommer.
\newblock High-resolution image synthesis with latent diffusion models.
\newblock In \emph{Proceedings of the IEEE/CVF conference on computer vision and pattern recognition}, pages 10684--10695, 2022.

\bibitem[Roy and Etemad(2024)]{roy2023consistency}
Shuvendu Roy and Ali Etemad.
\newblock Consistency-guided prompt learning for vision-language models.
\newblock In \emph{The Twelfth International Conference on Learning Representations}, 2024.

\bibitem[Soomro et~al.(2012)Soomro, Zamir, and Shah]{soomro2012ucf101}
Khurram Soomro, Amir~Roshan Zamir, and Mubarak Shah.
\newblock Ucf101: A dataset of 101 human actions classes from videos in the wild.
\newblock \emph{arXiv preprint arXiv:1212.0402}, 2012.

\bibitem[Villani et~al.(2009)]{villani2009optimal}
C{\'e}dric Villani et~al.
\newblock \emph{Optimal transport: old and new}.
\newblock Springer, 2009.

\bibitem[Wang et~al.(2019)Wang, Ge, Lipton, and Xing]{wang2019learning}
Haohan Wang, Songwei Ge, Zachary Lipton, and Eric~P Xing.
\newblock Learning robust global representations by penalizing local predictive power.
\newblock In \emph{NeurIPS}, 2019.

\bibitem[Wu et~al.(2025)Wu, Zhang, Li, Chen, Liang, Yang, and Li]{wu2025cascade}
Ge Wu, Xin Zhang, Zheng Li, Zhaowei Chen, Jiajun Liang, Jian Yang, and Xiang Li.
\newblock Cascade prompt learning for vision-language model adaptation.
\newblock In \emph{European Conference on Computer Vision}, pages 304--321. Springer, 2025.

\bibitem[Xiao et~al.(2010)Xiao, Hays, Ehinger, Oliva, and Torralba]{xiao2010sun}
Jianxiong Xiao, James Hays, Krista~A Ehinger, Aude Oliva, and Antonio Torralba.
\newblock Sun database: Large-scale scene recognition from abbey to zoo.
\newblock In \emph{CVPR}, pages 3485--3492. IEEE, 2010.

\bibitem[Yang et~al.(2024)Yang, Zhang, Wang, and Xie]{yang2024mma}
Lingxiao Yang, Ru-Yuan Zhang, Yanchen Wang, and Xiaohua Xie.
\newblock Mma: Multi-modal adapter for vision-language models.
\newblock In \emph{Proceedings of the IEEE/CVF Conference on Computer Vision and Pattern Recognition}, pages 23826--23837, 2024.

\bibitem[Yao et~al.(2024)Yao, Zhang, and Xu]{yao2024tcp}
Hantao Yao, Rui Zhang, and Changsheng Xu.
\newblock Tcp: Textual-based class-aware prompt tuning for visual-language model.
\newblock In \emph{Proceedings of the IEEE/CVF Conference on Computer Vision and Pattern Recognition}, pages 23438--23448, 2024.

\bibitem[Zhang et~al.(2022)Zhang, Zhang, Fang, Gao, Li, Dai, Qiao, and Li]{zhang2022tip}
Renrui Zhang, Wei Zhang, Rongyao Fang, Peng Gao, Kunchang Li, Jifeng Dai, Yu Qiao, and Hongsheng Li.
\newblock Tip-adapter: Training-free adaption of clip for few-shot classification.
\newblock In \emph{European conference on computer vision}, pages 493--510. Springer, 2022.

\bibitem[Zhou et~al.(2022{\natexlab{a}})Zhou, Yang, Loy, and Liu]{zhou2022conditional}
Kaiyang Zhou, Jingkang Yang, Chen~Change Loy, and Ziwei Liu.
\newblock Conditional prompt learning for vision-language models.
\newblock In \emph{CVPR}, pages 16816--16825, 2022{\natexlab{a}}.

\bibitem[Zhou et~al.(2022{\natexlab{b}})Zhou, Yang, Loy, and Liu]{zhou2022learning}
Kaiyang Zhou, Jingkang Yang, Chen~Change Loy, and Ziwei Liu.
\newblock Learning to prompt for vision-language models.
\newblock \emph{IJCV}, 130\penalty0 (9):\penalty0 2337--2348, 2022{\natexlab{b}}.

\bibitem[Zhu et~al.(2023)Zhu, Niu, Han, Wu, and Zhang]{zhu2023prompt}
Beier Zhu, Yulei Niu, Yucheng Han, Yue Wu, and Hanwang Zhang.
\newblock Prompt-aligned gradient for prompt tuning.
\newblock In \emph{Proceedings of the IEEE/CVF International Conference on Computer Vision}, pages 15659--15669, 2023.

\bibitem[Zhu et~al.(2024)Zhu, Ji, Zhao, Wu, and Wang]{zhu2024awt}
Yuhan Zhu, Yuyang Ji, Zhiyu Zhao, Gangshan Wu, and Limin Wang.
\newblock Awt: Transferring vision-language models via augmentation, weighting, and transportation.
\newblock \emph{arXiv preprint arXiv:2407.04603}, 2024.

\end{thebibliography}
}

\appendix

\renewcommand{\thefigure}{S\arabic{figure}}
\renewcommand{\thetable}{S\arabic{table}}

\section{Implementation Detail}\label{appendix:Implementation}

\begin{figure*}
    \centering
    \includegraphics[width=0.85\linewidth]{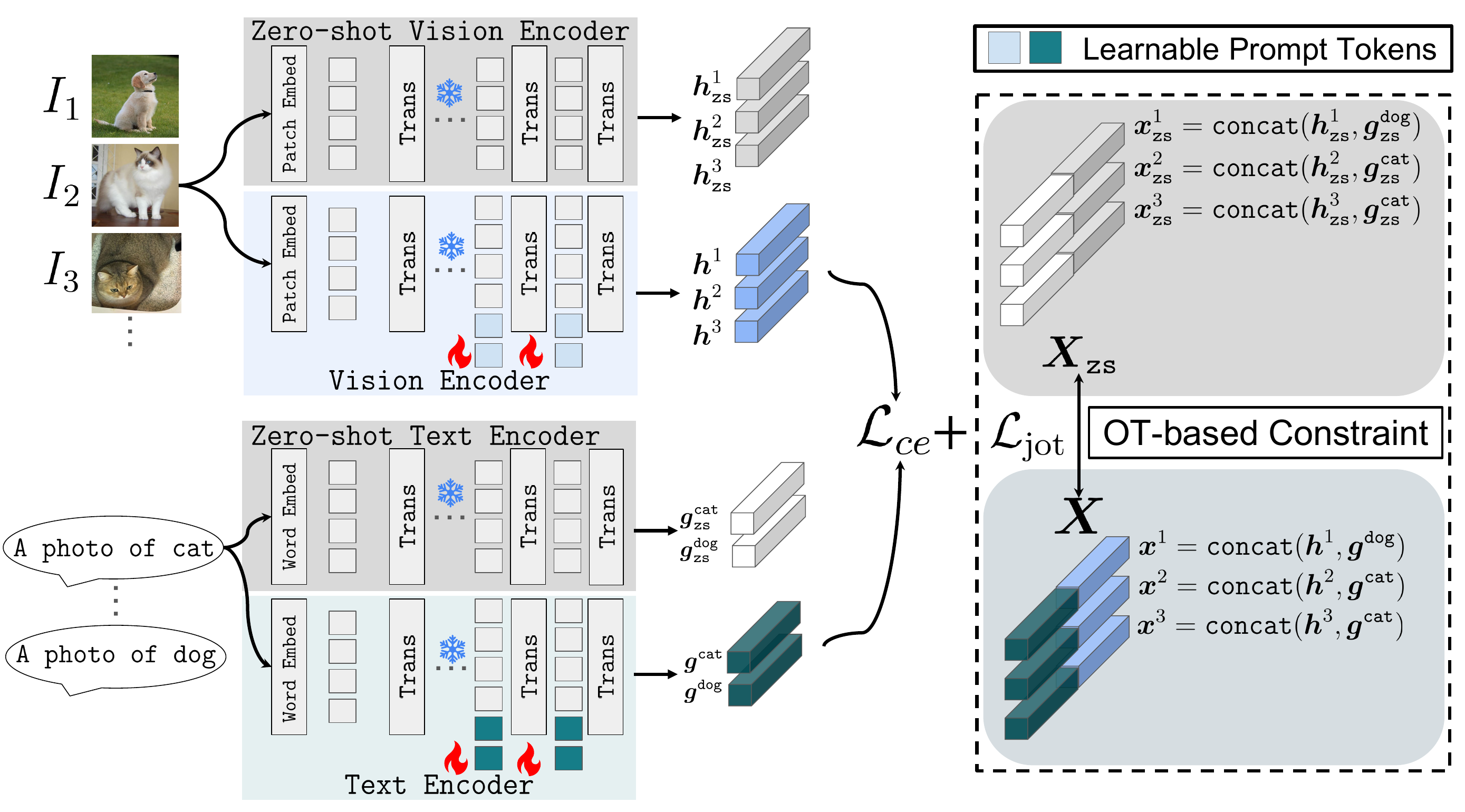}
    \vspace{-0.1in}
    \caption{The overview of our proposed framework. Only the prompt tokens are trainable, and the rest of the weights in both the zero-shot encoders and the adapted encoders are frozen. Despite the cross-entropy $\mathcal{L}_{ce}$ adopted, we also adopt the proposed joint optimal transport loss $\mathcal{L}_{\mathrm{jot}}$ between joint zero-short representation and adapted representation to constrain the model.}
    \label{fig:overview}
\end{figure*}

We train PromptOT for 30 epochs for the Base-to-Novel Generalization benchmark and 2 epochs for the remaining three benchmark settings, respectively. The respective epochs are fixed across all datasets. All experiments are run on a node of the cluster with one V100 16GB NVIDIA GPU. 
Following a deep prompting strategy, we apply prompts to all transformer blocks for the first benchmark and the first three transformer blocks for the remaining benchmarks, and we use four text prompts and four vision prompts.
Following previous works~\cite{khattak2023self}, in For Base-to-Novel Generalization benchmark, we set the first half of classes as base classes and the remaining classes as novel classes. Models are only trained on base classes. The accuracy is reported based on the evaluation of base classes and novel classes, respectively.
The overview of this framework is shown in Fig. \ref{fig:overview}.

\subsection{Datasets}
We used the following datasets in our experiments:

\noindent\textbf{Exps. 1 and 2.} ImageNet~\cite{deng2009imagenet}, Caltech101~\cite{fei2004learning}, OxfordPets~\cite{parkhi2012cats}, StanfordCars~\cite{krause20133d}, Flowers102~\cite{nilsback2008automated}, Food101~\cite{bossard2014food}, and FGVCAircraft~\cite{maji2013fine}, SUN397~\cite{xiao2010sun}, UCF101~\cite{soomro2012ucf101}, DTD~\cite{cimpoi2014describing}, and EuroSAT~\cite{helber2019eurosat}. 

\noindent\textbf{Exp. 3.} ImageNet~\cite{deng2009imagenet} as a source dataset and use ImageNet-A~\cite{hendrycks2021natural}, ImageNet-R~\cite{hendrycks2021many}, ImageNet-Sketch~\cite{wang2019learning} and ImageNetV2~\cite{recht2019imagenet}.

\subsection{Separate OT}

The way of applying \textbf{separate OT} is similar to SRC loss. Suppose the adapted vision features are presented as $\boldsymbol{H}=\{\boldsymbol{h}^i\}_{i=1}^{n}$ and adapted text features are presented as $\boldsymbol{G}=\{\boldsymbol{g}^i\}_{i=1}^{C}$. Again, $n$ and $C$ denote the number of samples and number of classes, respectively. Similarly, we denote the zero-shot vision features and text features as $\boldsymbol{H}_{zs}$ and $\boldsymbol{H}_{zs}$, respectively. The Separate OT is denoted as:
\begin{align}
    \mathcal{L}_{\texttt{SOT}} =   \mathcal{L}_{\texttt{OT}}(\boldsymbol{H},\boldsymbol{H}_{zs})+\mathcal{L}_{\texttt{OT}}(\boldsymbol{G},\boldsymbol{G}_{zs}).
\end{align}



\subsection{Reproducibility}
The code and model weights will be publicly released upon acceptance. \textbf{In the supplementary zip file, we provide the key code for the configuration and implementation of our proposed method.}

Our code is developed based on \url{https://github.com/muzairkhattak/PromptSRC}.

\section{Wilcoxon signed-rank test}
We perform a Wilcoxon signed-rank test~\cite{demvsar2006statistical} to compare our method with PromptSRC, the previous state-of-the-art, using harmonic mean accuracy as the evaluation metric. As shown in Table~\ref{tab:Wilcoxon}, our method achieves statistically significant improvements over PromptSRC (p-value < 0.05).

\begin{table}[]
\caption{Wilcoxon signed-rank test on Exp 1.}
\centering
\label{tab:Wilcoxon}
\resizebox{0.3\textwidth}{!}{%
\begin{tabular}{cc}\toprule
        & PromptSRC  VS Prompt-OT \\ \midrule
P-value & 0.016  \\ \bottomrule                
\end{tabular}%
}
\end{table}

\section{Limitation}\label{appendix:Limitation}
While our proposed PromptOT framework demonstrates strong performance across various benchmarks and introduces theoretically grounded improvements over prior prompt learning approaches, we acknowledge certain limitations that open avenues for future exploration:

First, the OT-based regularization introduces additional computational overhead compared to simpler constraints (e.g., point-wise L2 losses). Although we adopt mini-batch OT solvers to keep training efficient, the method still requires access to additional computational resources.

Second, while our approach avoids relying on external augmentations or auxiliary models (e.g., large LLMs), this also means it may not fully leverage certain recent advances in generative or data-enhancement techniques. Integrating such components in a resource-efficient manner could further boost performance.

\end{document}